\documentclass[10pt,twocolumn,letterpaper]{article}

\usepackage{cvpr}
\usepackage{times}
\usepackage{epsfig}
\usepackage{graphicx}
\usepackage{amsmath}
\usepackage{amssymb}
\usepackage[utf8]{inputenc}
\usepackage[nolist,nohyperlinks]{acronym}
\usepackage{booktabs}
\usepackage{multirow}
\usepackage{tabularx}
\usepackage{todonotes}
\usepackage{threeparttable}

\usepackage[pagebackref=true,breaklinks=true,letterpaper=true,colorlinks,bookmarks=false]{hyperref}

\begin{document}
\begin{acronym}
    \acro{FR}{face recognition}
    \acro{ABC}{automated border control}
    \acro{MAD}{morphing attack detection}
    \acro{GAN}{generative adversarial network}
    \acro{CNN}{convolutional neural network}
    \acro{FRLL}{Face Research Lab London}
    \acro{FRGC}{Face Recognition Grand Challenge}
    \acro{FFHQ}{Flicker-Faces HQ}
    \acro{GMM}{Gaussian Mixture Model}
    \acro{SVM}{Support Vector Machine}
    \acro{D-EER}{Detection Equal Error Rate}
    \acro{PCA}{Principal Component Analysis}
    \acro{LBP}{Local Binary Pattern}
    \acro{PRNU}{Photo Response Non-Uniformity}
\end{acronym}

\title{\vspace{-1cm}Evaluating the Effectiveness of Attack-Agnostic Features for Morphing Attack Detection}

\author{\large Laurent Colbois\textsuperscript{1,2} and Sébastien Marcel\textsuperscript{1,2}\\%
\textsuperscript{1} \small{Idiap Research Institute, Switzerland}\\%
\textsuperscript{2} \small{Université de Lausanne, Switzerland}\\%
{\tt\small \{laurent.colbois, sebastien.marcel\}@idiap.ch}
}
\maketitle
\thispagestyle{empty}

\begin{abstract}
    Morphing attacks have diversified significantly over the past years, with new methods based on generative adversarial networks (GANs) and diffusion models posing substantial threats to face recognition systems. Recent research has demonstrated the effectiveness of features extracted from large vision models pretrained on bonafide data only (attack-agnostic features) for detecting deep generative images. Building on this, we investigate the potential of these image representations for morphing attack detection (MAD). We develop supervised detectors by training a simple binary linear SVM on the extracted features and one-class detectors by modeling the distribution of bonafide features with a Gaussian Mixture Model (GMM). Our method is evaluated across a comprehensive set of attacks and various scenarios, including generalization to unseen attacks, different source datasets, and print-scan data. Our results indicate that attack-agnostic features can effectively detect morphing attacks, outperforming traditional supervised and one-class detectors from the literature in most scenarios. Additionally, we provide insights into the strengths and limitations of each considered representation and discuss potential future research directions to further enhance the robustness and generalizability of our approach.
 \end{abstract}

\section{Introduction}

\begin{figure}
\centering
\includegraphics[width=0.9\columnwidth]{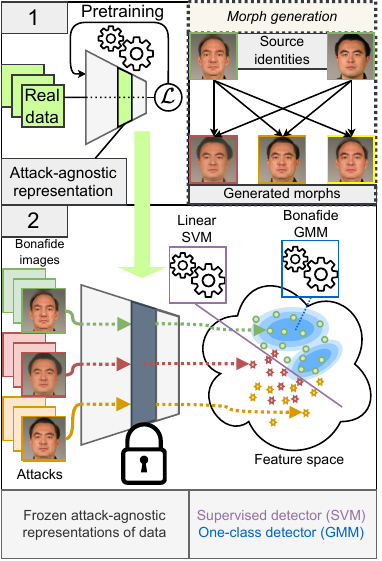}
\caption{We tackle the problem of \ac{MAD} using pretrained attack-agnostic extractors.
Morph generation: we generate morphs using a variety of algorithms (landmark-based, \ac{GAN}-based, and diffusion-based). Stage 1: the attack-agnostic extractor is a large vision model trained on real images for a pretext task. We reuse it to summarize any image by extracting an internal representation as the feature vector. Stage 2: features are extracted for bonafide images and face morphs. We train a supervised morphing attack detector as a linear SVM on top of this features space. We train a one-class detector by modeling the distribution of bonafide features with a \ac{GMM}, then using the likelihood of incoming samples as the discriminative score.}
\label{fig:methodology}
\end{figure}

Morphing attacks pose a significant threat to \ac{FR} systems. These attacks involve creating a composite passport image that merges facial features from two distinct source identities. This manipulated image is then submitted to governmental services for passport applications, a process still allowed in several European countries where applicants can provide their own photographs. In successful morphing attacks, both contributing individuals can then authenticate against the altered image, enabling them to share a single passport. This undermines the security and effectiveness of \ac{ABC} systems.

Historically, morphing attacks were primarily generated using straightforward image processing algorithms. These methods, typically relying on facial landmark warping, alignment, and pixel averaging of the source images, are known as \textbf{landmark-based} morphing techniques \cite{ferraraMagicPassport2014}. The vulnerability of existing \ac{FR} systems to such attacks has been well documented, prompting extensive research into morphing attack detection (\ac{MAD}) methods. 

More recently, advances in generative artificial intelligence have introduced fundamentally different morphing algorithms, particularly those based on \acp{GAN} \cite{venkateshCanGANGenerated2020, sarkarAreGANbasedMorphs2022, zhangMIPGANGeneratingStrong2021} and on diffusion models \cite{damerMorDIFFRecognitionVulnerability2023}. Resulting morphs are referred to as \textbf{deep morphs}, as they leverage deep neural networks for their creation. Initially less effective than landmark-based morphs, deep morphs have rapidly improved, and now also pose a significant concern as they reach real-world applicability.

Detection of deep morphs has often been approached by adapting existing methods designed for landmark-based morphs, such as incorporating deep morphs into the training datasets for data-driven detectors. We propose to consider the reverse perspective: treating \ac{MAD} as a \emph{deepfake} detection problem, specifically focusing on detecting deep synthetic images. With respect to typical deepfake detection, one must then address two additional challenges: keeping the ability to handle the fundamentally different nature of landmark-based morphs, and ensuring robustness against print-scan post-processing—a degradation not typically considered in deepfake detection literature.

Recent advancements in deepfake detection have demonstrated the unexpected effectiveness of using internal features from large vision models trained exclusively on real data. These features, which are then \textbf{attack-agnostic}, can be used in conjunction with simple downstream classifiers to perform detection. Notably, features extracted using pretrained CLIP models, originally trained for image-caption alignment, have shown promise in previous studies \cite{cozzolinoRaisingBarAIgenerated2023, ojhaUniversalFakeImage2023}.

This study focuses on evaluating the applicability of attack-agnostic features for \ac{MAD}. Specifically:
\begin{itemize}
\item We develop and evaluate \ac{MAD} systems using simple probe classifiers trained on attack-agnostic feature representations.
\item We develop and evaluate \ac{MAD} systems based on one-class modeling of the bona fide class, and detecting morphs as out-of-distribution samples, an approach which is enabled by the use of attack-agnostic representations.
\item We compare our methodology against traditional supervised \ac{CNN} training, through extensive experiments involving three different datasets and five types of morphing attacks spanning three categories: landmark-based, \ac{GAN}-based, and diffusion-based. Our evaluation includes a variety of scenarios, focusing on the generalization capabilities across different families of attacks, across source datasets, and across domains (digital to print-scan).
\end{itemize}

Source code for regenerating the morphs and reproduce the results is released publicly.\footnote{\url{https://gitlab.idiap.ch/bob/bob.paper.ijcb2024_agnostic_features_mad}}

\section{Related work}

\Acl{MAD} systems can be broadly categorized into single \ac{MAD} and differential \ac{MAD}. Single \ac{MAD} aims to assess the authenticity of a single image, such as a registered passport picture, while differential MAD also exploits probe information, such as the live-captured image of the passport holder at the Automated Border Control (ABC) gate. We focus here on single \ac{MAD} which our work is concerned with.

\Ac{MAD} systems can be categorized into those using handcrafted features and those using deep features \cite{scherhagFaceMorphingAttack2022}. Handcrafted features typically rely on texture cues (e.g. \acp{LBP}) or image forensic cues (e.g. frequency content, \ac{PRNU}). Deep features, on the other hand, are learned in a data-driven manner by training a neural network (usually a \ac{CNN}) on examples of bonafide and morphed images. 

A significant portion of research has focused on landmark-based morphs, with somewhat more limited attention given to \ac{GAN}-based morphs and almost none to the more recent diffusion-based morphs, such as those introduced in \cite{damerMorDIFFRecognitionVulnerability2023}. Common benchmark datasets, such as the NIST FATE MORPH \cite{FaceAnalysisTechnology} and the SOTAMD dataset \cite{rajaMorphingAttackDetectionDatabase2021}, include only a single type of deep morph (\ac{GAN}-based) or none at all. Similarly, the largest available dataset, SMDD \cite{damerPrivacyfriendlySyntheticData2022a}, based on synthetic identities, includes only a single landmark-based morphing attack. In practice, handcrafted features developed for landmark-based MAD are not particularly effective for deep morphs, as demonstrated in \cite{tapiaFaceFeatureVisualisation2023}. The effectiveness of deep features is strongly dependent on the training data, and generalization from a training dataset containing only landmark-based morphs to one containing deep morphs is not guaranteed, as observed in \cite{damerMorDIFFRecognitionVulnerability2023}.

Notable exceptions include two works that approach MAD as an anomaly detection problem. Both design an image-reconstruction network that aims to degrade then reconstruct bonafide input images. This process is done through an autoencoder in \cite{fangUnsupervisedFaceMorphing2022}, and by a noise-denoise process in \cite{ivanovskaFaceMorphingAttack2023} using diffusion models. They then observe that the reconstruction error differs between bonafide images and morphs, although it is \emph{lower} for morphs in \cite{fangUnsupervisedFaceMorphing2022} but \emph{higher} in \cite{ivanovskaFaceMorphingAttack2023}. The reconstruction error is thus discriminative for detection purposes. One main advantage of such approaches is that they are one-class, relying only on bonafide data and not on specific attacks in the training set, making them less prone to bias towards a specific family of morphing methods. However, evaluation on diffusion morphs, for example, is not provided in these works.

Finally, \cite{colboisDetectionMorphingAttacks2022} demonstrates that generic pre-existing GAN-image detectors are quite effective out-of-the-box for detecting GAN-based morphs in the digital domain. This suggests potential in leveraging methodologies from deep synthetic image detection research and applying them to \ac{MAD}. The two main additional challenges are handling landmark-based morphs, which are of a different nature, and dealing with the print-scan domain. Recent progress in synthetic image detection, as shown in \cite{ojhaUniversalFakeImage2023} and \cite{cozzolinoRaisingBarAIgenerated2023}, indicates that internal representations from existing large vision models, pretrained on auxiliary tasks and real data only (hence, \textbf{attack-agnostic}), can be surprisingly effective for synthetic image detection by training a simple downstream classifier on top of extracted features. Similarly to one-class approaches, this method is less prone to overspecialize for a family of morphing algorithms, given that the selected representations are based only on bonafide data.

The core goal of our work is to carefully examine the applicability of attack-agnostic features in the context of \ac{MAD}, particularly with the inclusion of landmark-based morphs and print-scan data.

\section{Methodology}
\subsection{Morph Datasets}
\begin{figure*}[t]
	\centering
	\begin{tikzpicture}
		\def\imagewidth{0.12\linewidth}
		\def\imagewspace{0.13\textwidth}
		\def\imagehspace{-0.13\textwidth}
		\def\texthspace{13pt}
		\node[anchor=north west, inner sep=0] (frgc0-source1) at (0,0) {\includegraphics[width=\imagewidth]{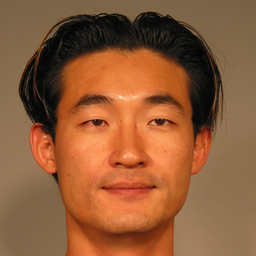}};
		\node[anchor=north] at (frgc0-source1.north) [yshift=\texthspace]  {Source 1};
		\node[anchor=north west, inner sep=0] (frgc0-landmark-complete) at (\imagewspace,0)
		{\includegraphics[width=\imagewidth]{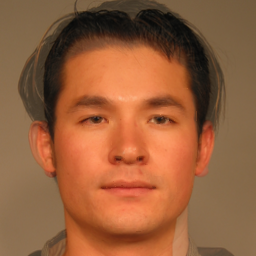}};
		\node[anchor=north] at (frgc0-landmark-complete.north) [yshift=\texthspace]  {LB-Complete};
		\node[anchor=north west, inner sep=0] (frgc0-landmark-combined) at (2*\imagewspace,0) {\includegraphics[width=\imagewidth]{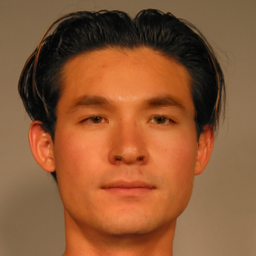}};
		\node[anchor=north] at (frgc0-landmark-combined.north) [yshift=\texthspace]  {LB-Combined};
		\node[anchor=north west, inner sep=0] (frgc0-stylegan-interp-W) at (3*\imagewspace,0) {\includegraphics[width=\imagewidth]{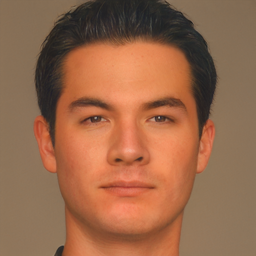}};
		\node[anchor=north] at (frgc0-stylegan-interp-W.north) [yshift=\texthspace]  {SG2-W};
		\node[anchor=north west, inner sep=0] (frgc0-stylegan-interp-W+) at (4*\imagewspace,0) {\includegraphics[width=\imagewidth]{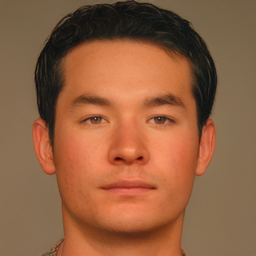}};
		\node[anchor=north] at (frgc0-stylegan-interp-W+.north) [yshift=\texthspace]  {SG2-W+};
		\node[anchor=north west, inner sep=0] (frgc0-diffae) at (5*\imagewspace,0) {\includegraphics[width=\imagewidth]{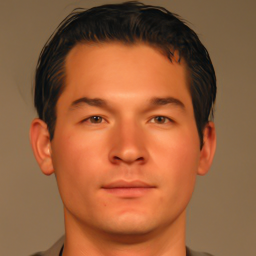}};
		\node[anchor=north] at (frgc0-diffae.north) [yshift=\texthspace]  {MorDIFF};
		\node[anchor=north west, inner sep=0] (frgc0-source2) at (6*\imagewspace,0)
		{\includegraphics[width=\imagewidth]{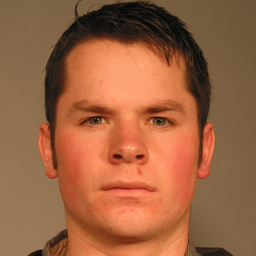}};
		\node[anchor=north] at (frgc0-source2.north) [yshift=\texthspace]  {Source 2};
		\node[anchor=north west, inner sep=0] (frll0-source1) at (0,\imagehspace) {\includegraphics[width=\imagewidth]{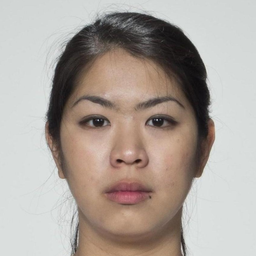}};
		\node[anchor=north west, inner sep=0] (frll0-landmark-complete) at (\imagewspace,\imagehspace)
		{\includegraphics[width=\imagewidth]{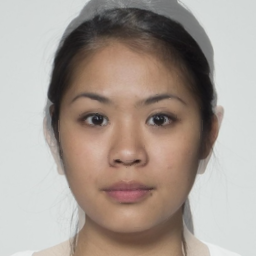}};
		\node[anchor=north west, inner sep=0] (frll0-landmark-combined) at (2*\imagewspace,\imagehspace) {\includegraphics[width=\imagewidth]{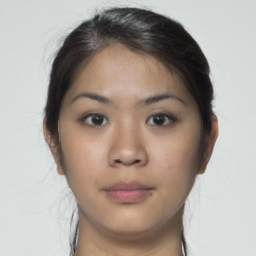}};
		\node[anchor=north west, inner sep=0] (frll0-stylegan-interp-W) at (3*\imagewspace,\imagehspace) {\includegraphics[width=\imagewidth]{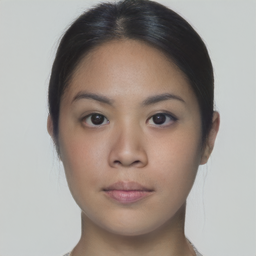}};
		\node[anchor=north west, inner sep=0] (frll0-stylegan-interp-W+) at (4*\imagewspace,\imagehspace) {\includegraphics[width=\imagewidth]{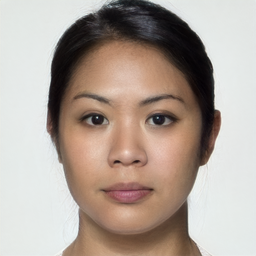}};
		\node[anchor=north west, inner sep=0] (frll0-diffae) at (5*\imagewspace,\imagehspace) {\includegraphics[width=\imagewidth]{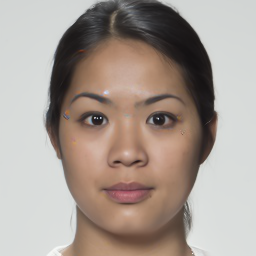}};
		\node[anchor=north west, inner sep=0] (frll0-source2) at (6*\imagewspace,\imagehspace)
		{\includegraphics[width=\imagewidth]{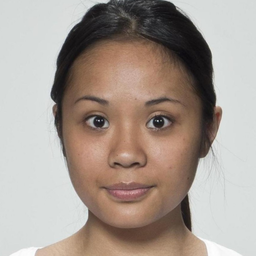}};
		\node[anchor=north west, inner sep=0] (ffhq0-source1) at (0,2*\imagehspace) {\includegraphics[width=\imagewidth]{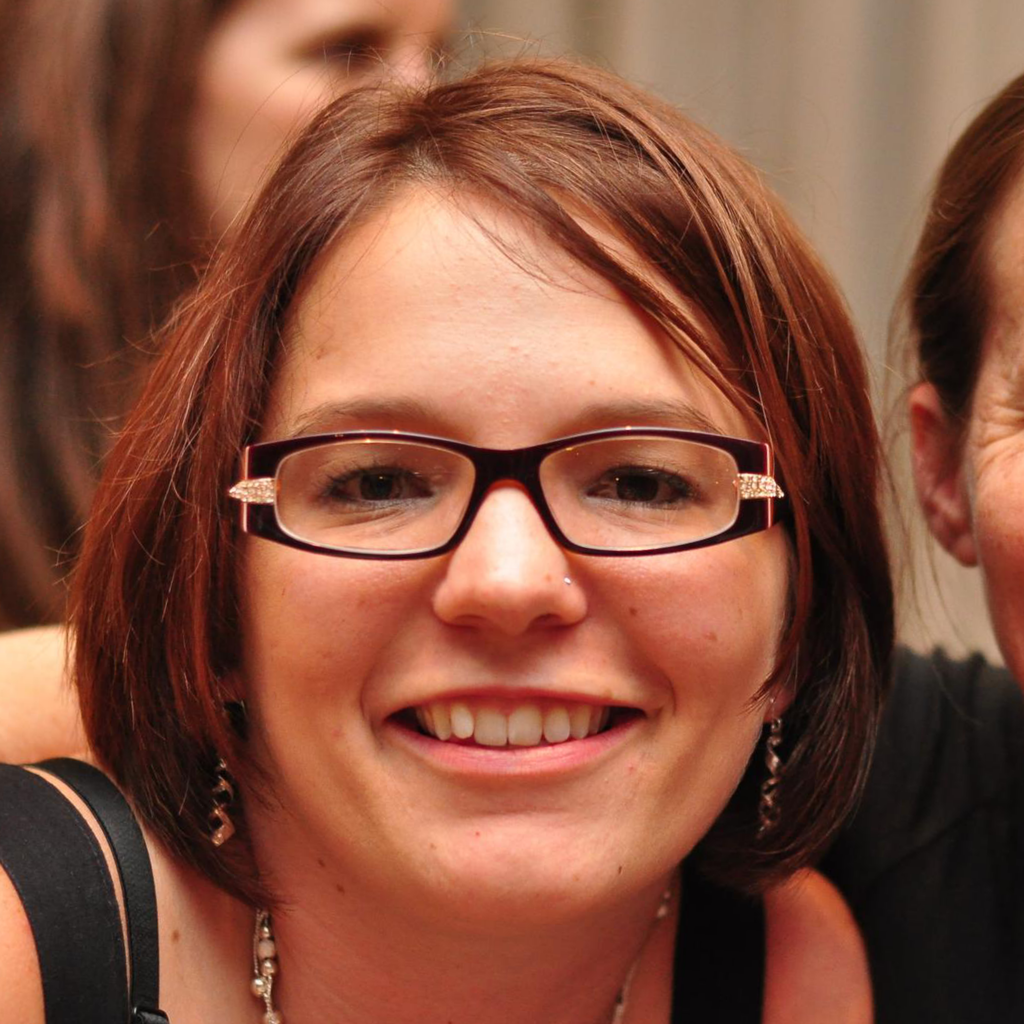}};
		\node[anchor=north west, inner sep=0] (ffhq0-landmark-complete) at (\imagewspace,2*\imagehspace)
		{\includegraphics[width=\imagewidth]{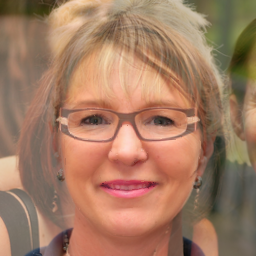}};
		\node[anchor=north west, inner sep=0] (ffhq0-landmark-combined) at (2*\imagewspace,2*\imagehspace) {\includegraphics[width=\imagewidth]{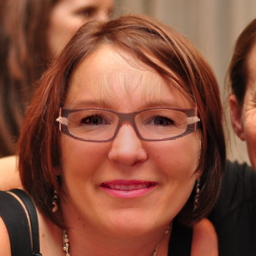}};
		\node[anchor=north west, inner sep=0] (ffhq0-stylegan-interp-W) at (3*\imagewspace,2*\imagehspace) {\includegraphics[width=\imagewidth]{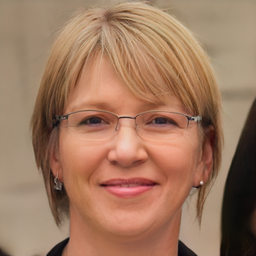}};
		\node[anchor=north west, inner sep=0] (ffhq0-stylegan-interp-W+) at (4*\imagewspace,2*\imagehspace) {\includegraphics[width=\imagewidth]{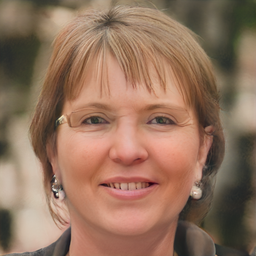}};
		\node[anchor=north west, inner sep=0] (ffhq0-diffae) at (5*\imagewspace,2*\imagehspace) {\includegraphics[width=\imagewidth]{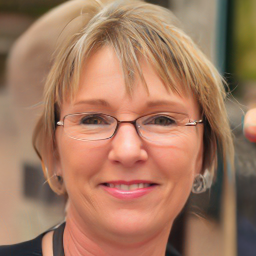}};
		\node[anchor=north west, inner sep=0] (ffhq0-source2) at (6*\imagewspace,2*\imagehspace)
		{\includegraphics[width=\imagewidth]{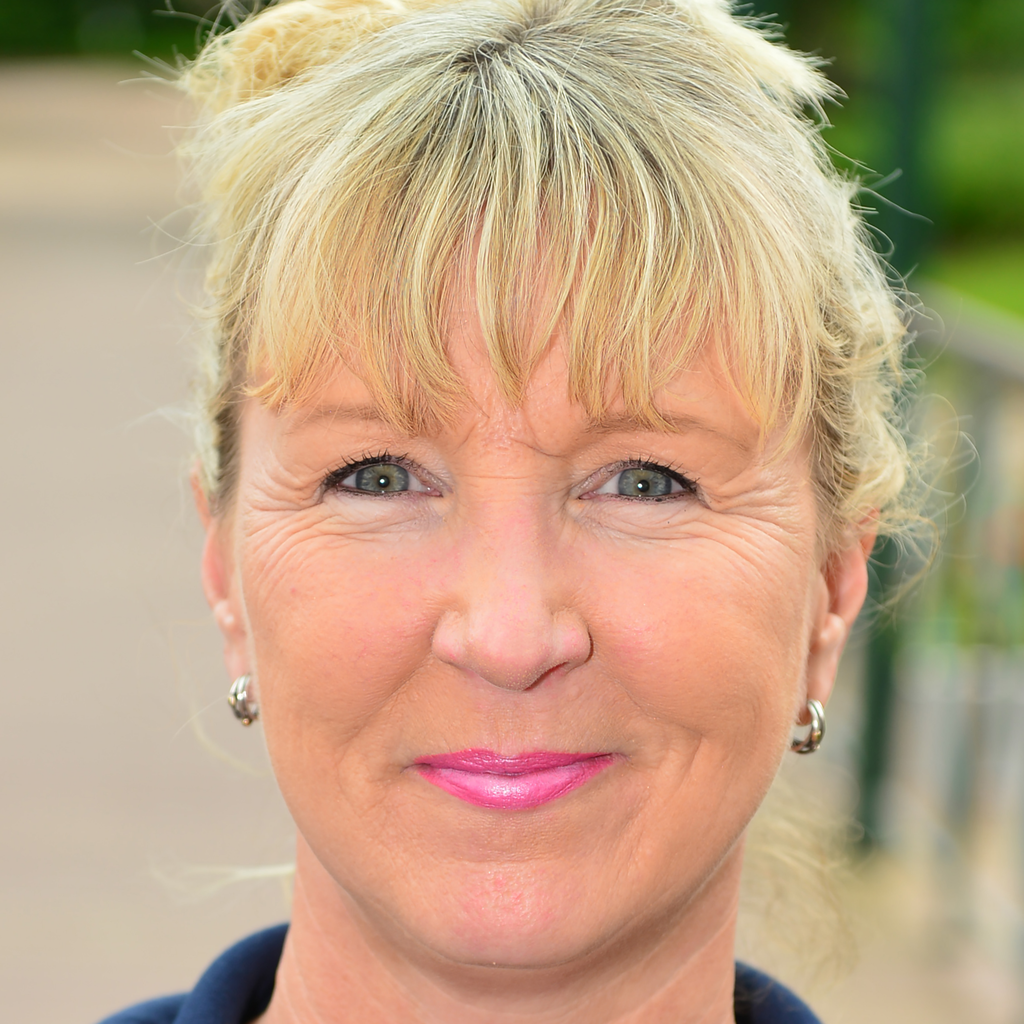}};

	\end{tikzpicture}
	\caption{Examples of generated morphs using as source dataset respectively FRGC (first row), FRLL (second row) and FFHQ (third row). The first and last column show the two real sources for which a morph must be created, and other columns show the results using each considered morphing algorithm.}
	\label{fig:morph_examples}
\end{figure*}
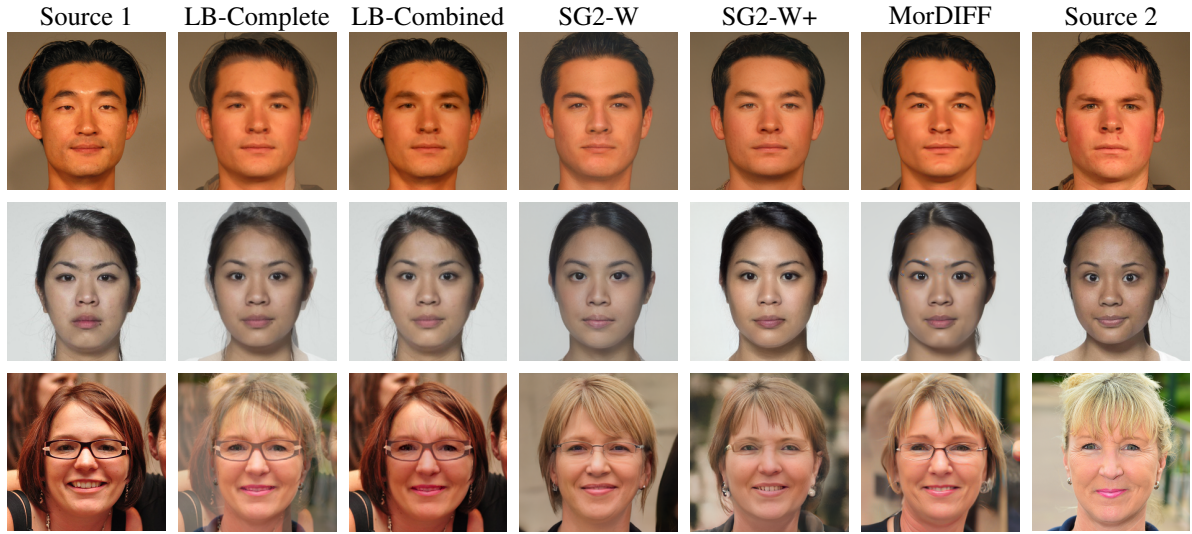

We create morphs using three distinct source datasets: the \ac{FRLL} dataset \cite{FaceResearchLab2017}, the \ac{FRGC} dataset \cite{phillipsOverviewFaceRecognition2005}, and the \ac{FFHQ} dataset \cite{karrasStyleBasedGeneratorArchitecture2019}. The \ac{FRLL} and \ac{FRGC} datasets are extensively utilized in prior research on face morphing due to their constrained facial images (frontal pose, neutral expression) with consistent backgrounds and illumination. These characteristics render them suitable for morph generation. In contrast, the \ac{FFHQ} dataset, collected from Flickr, exhibits greater diversity. We hypothesize that employing a more diverse source dataset is advantageous for our research objectives, particularly in studying cross-source dataset generalization and one-class modeling of the bonafide class.

For the \ac{FRLL} and \ac{FRGC} datasets, we select identity pairs for morph creation following previous research works \cite{neubertExtendedStirTraceBenchmarking2018} and \cite{zhangMIPGANGeneratingStrong2021}, respectively. This results in 1,140 pairings for \ac{FRLL} and 2,521 pairings for \ac{FRGC}. For the \ac{FFHQ} dataset, we initially select 10,000 images from the original dataset of 70,000 images, focusing on those with the most frontal poses, which are then randomly paired to form 5,000 morphing pairs. While this process might yield some unrealistic morphs (e.g., morphs between different genders), it allows the creation of a large set of samples containing the relevant attack artifacts and showcasing high diversity. Hence, this set remains valuable for training \ac{MAD} systems.

Using consistent pairings, we generate morphs from these source datasets employing five different attack algorithms. These include two landmark-based algorithms (LB-Complete \cite{makrushinAutomaticGenerationDetection2017} and LB-Combined \cite{neubertExtendedStirTraceBenchmarking2018}), two GAN-based algorithms (SG2-W \cite{karrasStyleBasedGeneratorArchitecture2019} and SG2-W+ \cite{venkateshCanGANGenerated2020}), and one diffusion-based algorithm (MorDIFF \cite{damerMorDIFFRecognitionVulnerability2023}). Examples of the generated morphs are presented in Figure \ref{fig:morph_examples}.

For the bonafide sets, we use original images from the source datasets. For FRLL, we use the only available 204 frontal images, some of which have also been used as sources for morphing. Due to this low amount of bonafide images, we restrict the usage of FRLL to test purposes. For FRGC and FFHQ, we select bonafide images containing identities never used for morphing, with 11,532 and 10,000 images, respectively. We split both the bonafide sets and attack sets into training and test sets using an 80-20 ratio, ensuring that identities are disjoint between the training and test sets for bonafide images, and that pairs of identities are disjoint between the training and test sets for the attacks. The exact number of samples in each dataset is detailed in Table \ref{tab:data_stats}.
\begin{table}
    \centering
    \begin{threeparttable}
    \caption{Number of samples in each dataset and split. We indicate the number of attack samples \emph{per morphing algorithm}, i.e. the total number of attack samples used in experiments should be obtained by multiplying the provided value by the number of considered morphing algorithms.}
    \label{tab:data_stats}

    \small{
        \renewcommand{\arraystretch}{1.2}
        \begin{tabular}{@{}cccccc@{}}
            \toprule
                         & \multicolumn{2}{c}{\# bonafide} & \phantom{;} & \multicolumn{2}{c}{\# per attack}                \\
            \cmidrule{2-3} \cmidrule{5-6}
            Src. dataset & Train                           & Test        &                                   & Train & Test \\
            \midrule
            FRGC         & 9228                            & 2304        &                                   & 2014  & 507  \\
            FRLL         & -                               & 204         & -                                 &       & 1140 \\
            FFHQ         & 8000                            & 2000        &                                   & 4000  & 1000 \\
            \bottomrule
        \end{tabular}
    }
    \end{threeparttable}

\end{table}

\begin{table}
    \centering
    \begin{threeparttable}
    \caption{Available image sets. Attacks are grouped into higher level families indicated in the first row. Most attacks are available in digital format ($\circ$), some of them have their test set in print-scan format as well ($\bullet$). The FRGC-MIPGAN attack is used only for testing purpose in the print-scan domain ($\star$).}
    \label{tab:morph_sets}

    \small{
        \renewcommand{\arraystretch}{1.2}
        \begin{tabular}{@{}lcccccccccc@{}}
            \toprule
                 & \multicolumn{2}{c}{LB}   & \phantom{;}                                                                                          & \multicolumn{2}{c}{GAN}                                                           & \phantom{;}                                                                         & Diff                                                                                             & \phantom{;}                                                                            &                                                  & \\
                 \cmidrule{2-3} \cmidrule{5-6} \cmidrule{8-8}                 
                 & \rotatebox{90}{\footnotesize{\textbf{LB-Complete}}\cite{makrushinAutomaticGenerationDetection2017}} & \rotatebox{90}{  \footnotesize{\textbf{LB-Combined}}\cite{neubertExtendedStirTraceBenchmarking2018}} && \rotatebox{90}{ \footnotesize{\textbf{SG2-W}}\cite{sarkarAreGANbasedMorphs2022} } & \rotatebox{90}{  \footnotesize{\textbf{SG2-W+}}\cite{venkateshCanGANGenerated2020}} && \rotatebox{90}{  \footnotesize{\textbf{MorDIFF}}\cite{damerMorDIFFRecognitionVulnerability2023}} && \rotatebox{90}{ \footnotesize{\textbf{MIPGAN}}\cite{zhangMIPGANGeneratingStrong2021} } & \rotatebox{90}{\footnotesize{\textbf{Bonafide}}}   \\
            \midrule
            FRGC & $\circ$                                                                                             & $\bullet$                                                                                            && $\circ$                                                                           & $\circ$                                                                             && $\circ $                                                                                         && $\star$                                                                                & $\bullet$                                          \\
            FRLL & $\circ$                                                                                             & $\circ$                                                                                              && $\circ$                                                                           & $\circ$                                                                             && $\circ$                                                                                          &&                                                                                        & $\circ$                                            \\
            FFHQ & $\circ$                                                                                             & $\circ$                                                                                              && $\circ$                                                                           & $\circ$                                                                             && $\circ$                                                                                          &&                                                                                        & $\circ$                                            \\
            \bottomrule
        \end{tabular}
    }
    \end{threeparttable}

\end{table}

Additionally, we create a "real-world" test dataset by printing and scanning a subset of images. Specifically, the bonafide test samples from FRGC, morph test samples created using FRGC with the LB-Combined and MorDIFF algorithms, and an additional set of FRGC morphs created using another unseen algorithm, MIPGAN \cite{zhangMIPGANGeneratingStrong2021}. This simulates a challenging scenario where we must generalize from the digital to the print-scan domain, and towards unseen attacks. The morphs are printed at a size of 35mm x 35mm then rescanned at a resolution of 300 DPI, using a \textit{Kyocera TASKalfa 2554ci} (laser printer + scanner).
As preprocessing, all images are cropped to 256x256 pixels while ensuring consistent landmark alignement.

Available image sets are summarized in Table \ref{tab:morph_sets}. For the experiments, we regroup attacks into higher level families, respectively landmark-based (LB), GAN-based (GAN), and diffusion-based (Diff).

\subsection{Evaluation Scenarios}

We aim to evaluate the performance of \ac{MAD} systems in various settings, with a focus on assessing generalization capability across unseen attack families (LB, GAN, or Diff), unseen source datasets, and different domains (digital to print-scan). Additionally, we seek to evaluate the performance of one-class detectors trained solely on bonafide data. The following evaluation scenarios are considered:

\begin{enumerate}
\item{\textbf{Baseline} : the detector is trained and tested on digital bonafide and morph samples from the same source dataset (FRGC or FFHQ), with all families of attacks seen during training.}
\item \textbf{Generalization to unseen attacks} : unlike the baseline, the detector is trained using only a single family of attacks (LB, GAN, or Diff.) and tested on the other two families.
\item \textbf{Generalization to different source datasets} : unlike the baseline, the detector is tested on bonafide and morph samples from an unseen source dataset, specifically FRLL.
\item \textbf{Generalization to print-scan data} : unlike the baseline, the detector is tested on print-scanned bonafide and morph samples. This scenario is evaluated only using FRGC, for which print-scanned data is available.
\item \textbf{One-Class Detection} : the detector is trained solely on bonafide samples and then tested on all attacks. For this setting, we restrict ourselves to a single source dataset and to the digital domain.
\end{enumerate}

The first four scenarios involve training the detector in a supervised manner as a binary classifier. In the last scenario, one-class detectors are achieved by modeling the statistical distribution of the features of bonafide samples, then using the likelihood score of incoming samples under the learned distribution as the discriminative score.
For both types of systems, performance is evaluated by reporting the \ac{D-EER} on the respective test sets.

\subsection{Models}

We consider two types of detection models. The first type, which is the focus of our study, involves training a simple downstream classifier on top of pretrained features extracted from an attack-agnostic vision model, i.e., a network trained solely on bonafide data for some auxiliary task (cf. Figure \ref{fig:methodology}). The second type is used for comparison purposes, and consist in fully training a convolutional neural network directly on image samples, either as a binary classifier (in the supervised setting) or as an autoencoder (in the one-class setting).

\subsubsection{Probed Attack-Agnostic Models}
We consider the following attack-agnostic feature extractors:

\begin{itemize}
    \item{\textbf{RN50-IN} \cite{paszkePyTorchImperativeStyle2019} : this baseline extractor is a ResNet50 network trained for image classification on ImageNet. We use the output of the penultimate layer before the image classification layer as the feature representation of images.}
    \item{\textbf{DINOv2} \cite{oquabDINOv2LearningRobust2023} : this extractor is trained in a self-supervised manner with the goal of learning general image representations. It has demonstrated effectiveness for a broad variety of downstream classification tasks and serves as a more sophisticated baseline compared to RN50-IN. We specifically use the `giant` variant, and use the learned general representation as feature vector.}
    \item{\textbf{CLIP} \cite{radfordLearningTransferableVisual2021} : this vision-language model is trained to represent matched image-caption pairs jointly in the same feature space. Despite being trained for a seemingly unrelated task, previous research \cite{ojhaUniversalFakeImage2023,cozzolinoRaisingBarAIgenerated2023} has shown that CLIP-extracted features showcase strong discriminative power to differentiate between bonafide and synthetic images. We use the L/14 variant as suggested by \cite{ojhaUniversalFakeImage2023}, and use the output of the vision encoder as feature vector.}
    \item {\textbf{AIM} \cite{el-noubyScalablePretrainingLarge2024} : this extractor is pretrained for auto-regressive image modeling, which involves decomposing images into ordered sequences of patches and predicting subsequent patches using only the context of previous patches. This auto-regressive objective is theoretically equivalent to learning the true underlying image distribution. Trained on a massive dataset of 12.8 billion images, AIM has the potential to approximate the distribution of "natural" images. Given that deep synthetic images typically exhibit salient statistical differences from bonafide ones \cite{cozzolinoRaisingBarAIgenerated2023}, we hypothesize that they might lie outside of the distribution learned by AIM. We use the 600M variant, and use the pool-averaged output of the trunk as the feature representation.}
    \item{\textbf{DNADet} \cite{yangDeepfakeNetworkArchitecture2022a} : this extractor is originally designed to improve the accuracy of source attribution for GAN-generated images. It is pretrained using real images for a task of patchwise contrastive learning of image transformations, where images undergo various degradations (e.g., blurring, JPEG compression) and are decomposed into patches. The model learns to represent patches subject to the same degradations close to each other, and patches subject to different degradations far apart, and additionally has to classify incoming patches based on their applied degradation. Given the already demonstrated efficacy of this pretraining in learning salient features to differentiate various GAN models, DNADet is a strong candidate for synthetic image detection, particularly as its pretraining data includes face images, making it also content-specific for our case. We use the output of the penultimate layer, right before the fully connected layer used for the classification, as the feature representation.}
\end{itemize}
For supervised modeling, we train a downstream linear probe on top of the extracted features, specifically a binary linear \ac{SVM}, preceded by a \ac{PCA} decomposition achieving 99\% of explained variance. This initial projection mitigates the challenges posed by the high dimensionality of certain feature spaces.

For one-class modeling, we fit the distribution of bonafide features using a \ac{GMM}, also preceded by \ac{PCA} decomposition achieving 99\% of explained variance. The log-likelihood of incoming samples under this statistical model is then used to distinguish between bonafide samples, which are expected to have high log-likelihood values, and attacks, which are expected to have low log-likelihood values. To determine the optimal number of components for the \ac{GMM} (ranging from 1 to 256), as well as the type of covariance matrix (diagonal or spherical), we perform 4-fold cross-validation on the training set. The validation set includes attack samples, and the \ac{D-EER} on the validation set is used as the selection criterion.

\subsubsection{Reference \ac{MAD} Models}
For the supervised detection setting, we use as comparative reference the MixFaceNet architecture, which has been employed in prior work as a backbone for training \ac{MAD} systems. The model is a \ac{CNN} trained as a binary classifier directly on image examples. We reproduce the backbone setup and training process as described in \cite{damerPrivacyfriendlySyntheticData2022a}, and reuse their provided code.\footnote{\url{https://github.com/naserdamer/SMDD-Synthetic-Face-Morphing-Attack-Detection-Development-dataset}}
For the one-class setting, we compare our models to the SPL-MAD model from \cite{fangUnsupervisedFaceMorphing2022}. The SPL-MAD model is trained as a convolutional autoencoder on the Casia-WebFace dataset (bonafide face images). At inference time, the authors observe that the reconstruction error is smaller for morphs than for bonafide images, and thus can be used as a discriminative score for detection, despite using only bonafide data at training time. We use the code and pretrained model provided by the authors\footnote{\url{https://github.com/meilfang/SPL-MAD}}.

\section{Results}

\paragraph{Baseline performance}
\begin{table}[t]
    \centering
    \begin{threeparttable}

        \caption{\textbf{Baseline.} \ac{D-EER} (\%) on the test split when all attacks are seen at training time. \textbf{Bold} values indicate setups where probed attack-agnostic models perform better than the MixFaceNet MAD reference. \underline{Underlined} values are the best performing models.}
        \label{tab:baseline}

        \small{
            \renewcommand{\arraystretch}{1.2}
            \begin{tabular}{@{}lccccccc@{}}
                \toprule
                Src dataset  & \multicolumn{3}{c}{FRGC} & \phantom{;}      & \multicolumn{3}{c}{FFHQ}                                                                                            \\
                Test on      & LB                       & GAN              & DIFF                     &                           & LB                        & GAN                       & DIFF \\
                \midrule
                Model        &                          &                  &                          &                           &                           &                           &      \\
                AIM          & \underline{0.00}         & \underline{0.00} & \underline{0.00}         && \textbf{\underline{0.20}} & \textbf{\underline{0.05}} & \textbf{\underline{0.05}}        \\
                CLIP         & \underline{0.00}         & \underline{0.00} & 0.13                     && \textbf{1.45}             & \textbf{0.40}             & \textbf{1.65}                    \\
                DNADet       & \underline{0.00}         & 0.04             & \underline{0.00}         && \textbf{5.70}             & 5.75                      & \textbf{6.10}                    \\
                DINOv2       & \underline{0.00}         & \underline{0.00} & \underline{0.00}         && \textbf{5.50}             & \textbf{2.45}             & \textbf{3.25}                    \\
                RN50-IN      & 0.04                     & 0.17             & \underline{0.00}         && 9.70                      & 7.25                      & \textbf{6.35}                    \\
                MixFaceNet * & \underline{0.00}         & \underline{0.00} & \underline{0.00}         && 6.70                      & 5.30                      & 7.05                             \\
                \bottomrule
            \end{tabular}
        }
    \end{threeparttable}

\end{table}

Table \ref{tab:baseline} shows results for the baseline scenario where both training and testing data come from the same source dataset in the digital domain, and all attacks are known during training. For attacks from a constrained dataset (\ac{FRGC}), all methods perform well, achieving nearly perfect separation between bonafide and attack samples regardless of the attack family. However, with a more diverse dataset (\ac{FFHQ}), performance declines for most methods, and differences become more evident. Here, our linear probes generally outperform the MixFaceNet detectors, with AIM and CLIP achieving the best results across all attack families.

\paragraph{Generalization to unseen attacks}
\begin{table*}[t]
    \centering
    \begin{threeparttable}
        \caption{\textbf{Unseen attacks generalization.} \ac{D-EER} (\%) on the test split when a single family of attacks is seen at training time. \textbf{Bold} values indicate setups where probed attack-agnostic models perform better than the MixFaceNet MAD reference. \underline{Underlined} values are the best performing models.}
        \label{tab:cross_attack}

        \small{
            \renewcommand{\arraystretch}{1.2}
            \begin{tabular}{@{}llccccccccccc@{}}
                \toprule
                                         & Train attacks & \multicolumn{3}{c}{LB}  & \phantom{;}  & \multicolumn{3}{c}{GAN}   & \phantom{;} & \multicolumn{3}{c}{Diff}                                                                                                                                                                          \\
                                         \cmidrule{3-5} \cmidrule{7-9} \cmidrule{11-13}
                                         & Test attacks  & LB                        & GAN                       & Diff              &        & LB                        & GAN                       & Diff               &       & LB                        & GAN                       & Diff                      \\
                \midrule
                Src. dataset             & Model         &                           &                           &     &&                      &                           &                           &                           &                           &                           &                           \\
                \multirow[c]{6}{*}{FRGC} & AIM           & \underline{0.00}          & \textbf{\underline{0.00}} & \textbf{\underline{0.00}} & & \textbf{\underline{0.22}} & \underline{0.00}          & \textbf{\underline{0.39}} && 33.81                     & 8.68                      & \underline{0.00}          \\
                                         & CLIP          & \underline{0.00}          & \textbf{\underline{0.00}} & 1.22                      && 5.21                      & \underline{0.00}          & 5.03                      && 4.34                      & \textbf{\underline{0.22}} & \underline{0.00}          \\
                                         & DNADet        & \underline{0.00}          & 2.91                      & \textbf{\underline{0.00}} && 9.77                      & \underline{0.00}          & \textbf{0.65}             && 1.39                      & 1.09                      & \underline{0.00}          \\
                                         & DINOv2        & \underline{0.00}          & 0.69                      & 1.13                      && 10.81                     & \underline{0.00}          & 5.90                      && 7.34                      & 2.78                      & \underline{0.00}          \\
                                         & RN50-IN       & 0.09                      & 6.03                      & \textbf{\underline{0.00}} && 11.50                     & 0.09                      & \textbf{0.74}             && 2.86                      & 4.86                      & \underline{0.00}          \\
                                         & MixFaceNet *  & \underline{0.00}          & 0.48                      & 0.13                      && 2.73                      & \underline{0.00}          & 1.87                      && \underline{0.95}          & 0.61                      & \underline{0.00}          \\
                \cline{1-2}
                \multirow[c]{6}{*}{FFHQ} & AIM           & \textbf{\underline{0.00}} & \textbf{11.20}            & \textbf{19.60}            && \textbf{12.90}            & \textbf{\underline{0.00}} & \textbf{11.90}            && 27.90                     & \textbf{13.00}            & \textbf{\underline{0.00}} \\
                                         & CLIP          & \textbf{1.25}             & \textbf{\underline{0.90}} & \textbf{\underline{8.30}} && \textbf{\underline{5.70}} & \textbf{\underline{0.00}} & \textbf{\underline{7.90}} && \textbf{\underline{7.75}} & \textbf{\underline{1.20}} & \textbf{0.30}             \\
                                         & DNADet        & \textbf{2.30}             & \textbf{17.05}            & \textbf{27.20}            && \textbf{16.15}            & 1.95                      & 41.45                     && \textbf{25.30}            & 26.35                     & \textbf{0.90}             \\
                                         & DINOv2        & 5.10                      & \textbf{8.25}             & \textbf{12.20}            && \textbf{20.75}            & \textbf{0.30}             & \textbf{17.85}            && \textbf{19.05}            & \textbf{10.15}            & \textbf{0.55}             \\
                                         & RN50-IN       & 7.75                      & 33.00                     & \textbf{17.65}            && 33.90                     & 2.45                      & 36.60                     && \textbf{21.90}            & 26.95                     & \textbf{2.25}             \\
                                         & MixFaceNet *  & 5.00                      & 18.10                     & 33.75                     && 24.05                     & 0.85                      & 34.55                     && 26.15                     & 26.15                     & 2.40                      \\
                \bottomrule
            \end{tabular}
        }
    \end{threeparttable}

\end{table*}

Table \ref{tab:cross_attack} presents results where only one attack family is known during training. For \ac{FRGC} attacks, AIM features perform best except when only diffusion attacks are known; in this case, MixFaceNet is superior for diffusion to landmark-based generalization, and CLIP is best for diffusion to GAN generalization. The DNADet probe shows comparable generalization from diffusion to both landmark-based and GAN attacks, though it performs slightly worse than MixFaceNet overall.

For \ac{FFHQ} attacks, CLIP probes consistently outperform AIM probes and MixFaceNet across all generalization scenarios. Our linear probing approach also frequently surpasses MixFaceNet.

\paragraph{Generalization to difference source datasets}
\begin{table}
    \centering
    \begin{threeparttable}
    \caption{\textbf{Source dataset generalization.} \ac{D-EER} (\%) on FRLL bona fide \& morph images when all attacks based on a \emph{different} source dataset are seen at training time. \textbf{Bold} values indicate setups where probed attack-agnostic models perform better than the MixFaceNet MAD reference. \underline{Underlined} values are the best performing models.}
    \label{tab:cross_src}
    \footnotesize{
        \renewcommand{\arraystretch}{1.2}
        \begin{tabular}{@{}lccccccc@{}}
            \toprule
            Train src. dataset & \multicolumn{3}{c}{FRGC}  & \phantom{;}               & \multicolumn{3}{c}{FFHQ}                                                                                         \\
            \cmidrule{2-4} \cmidrule{6-8}
            Test attacks & LB                        & GAN                       & DIFF                      &  & LB                        & GAN                       & DIFF                      \\
            \midrule
            Model        &                           &                           &                           &  &                           &                           &                           \\
            AIM          & \textbf{\underline{1.47}} & \textbf{23.53}            & \textbf{11.76}            &  & \textbf{\underline{0.00}} & \textbf{\underline{0.00}} & \textbf{\underline{0.00}} \\
            CLIP         & \textbf{6.86}             & \textbf{\underline{4.90}} & \textbf{7.84}             &  & 3.43                      & \textbf{0.49}             & \textbf{0.98}             \\
            DNADet       & \textbf{10.29}            & 35.78                     & 42.65                     &  & \textbf{\underline{0.00}} & \textbf{\underline{0.00}} & \textbf{\underline{0.00}} \\
            DINOv2       & \textbf{9.80}             & \textbf{8.33}             & \textbf{\underline{3.92}} &  & 15.20                     & 13.24                     & 5.88                      \\
            RN50-IN      & 13.24                     & 29.41                     & \textbf{19.61}            &  & \textbf{1.96}             & 38.73                     & \textbf{0.98}             \\
            MixFaceNet * & 12.75                     & 28.92                     & 20.10                     &  & 2.94                      & 11.76                     & 1.47                      \\
            \bottomrule
        \end{tabular}
    }
    \end{threeparttable}

\end{table}

Table \ref{tab:cross_src}  reports results when the source dataset differs between training and testing, focusing on the \ac{FRLL} dataset. The experiment highlights in particular the importance of source dataset diversity for effective generalization. Indeed, detectors trained on FFHQ attacks generally perform better on FRLL attacks than those trained on FRGC. This trend holds for both our linear probes and the MixFaceNet detector, with DINOv2 being a notable exception. Overall, linear probes typically outperform MixFaceNet. When trained on FFHQ attacks, AIM and DNADet probes achieve perfect separation between FRLL morphs and bonafide samples but perform poorly when trained on FRGC attacks. In this latter case, CLIP probes provide the most balanced performance for generalization to unseen datasets across all attack types.

\paragraph{Generalization to print-scan data}
\begin{table}
    \centering
    \begin{threeparttable}
    \caption{\textbf{Print-scan generalization.} \ac{D-EER} (\%) on test split when all digital attacks are seen at training time, but test attacks are in the print-scan domain. \textbf{Bold} values indicate setups where probed attack-agnostic models perform better than the MixFaceNet MAD reference. \underline{Underlined} values are the best performing models.}
    \label{tab:print_scan}

    \small{
        \renewcommand{\arraystretch}{1.2}

        \begin{tabular}{@{}lccc@{}}
            \toprule
            Src dataset & \multicolumn{3}{c}{FRGC} \\
            \cmidrule{2-4}
            Test on & LB-PS & MIPGAN-PS & DIFF-PS \\
            \midrule
            Model &  &  &  \\
            AIM & \textbf{4.77} & \textbf{32.47} & \textbf{30.12} \\
            CLIP & \textbf{\underline{3.99}} & \textbf{15.02} & \textbf{14.97} \\
            DNADet & \textbf{16.19} & 60.50 & 56.55 \\
            DINOv2 & \textbf{8.85} & \textbf{\underline{5.60}} & \textbf{\underline{7.51}} \\
            RN50-IN & \textbf{20.57} & \textbf{35.24} & \textbf{26.78} \\
            MixFaceNet * & 22.05 & 50.22 & 32.34 \\
            \bottomrule
            \end{tabular}
    }
        \end{threeparttable}

\end{table}
    
Table \ref{tab:print_scan} shows results when detectors trained on digital data are evaluated on print-scan data. This scenario is challenging because artifacts left by deep morph generators on generated samples are likely degraded during the print-scan process. Most detectors, which achieved perfect separation in the baseline protocol, show significant performance drops on print-scan data (notably LB-PS and Diff-PS, even though they contain attacks whose digital counterpart has been seen during training). The MIPGAN-PS attacks are particularly challenging due to being totally unseen during training. Neverthesless, our linear probes still generally outperform MixFaceNet, with DINOv2 features being the most effective, followed by CLIP.

\paragraph{One-class detector}

\begin{table}
    \centering
    \begin{threeparttable}
    \caption{\textbf{One-class model.} \ac{D-EER} (\%) on the test split  when only bona fide sample are seen at training time. We compare to the SPL-MAD model from \cite{fangUnsupervisedFaceMorphing2022}. \textbf{Bold} values indicate setups where probed attack-agnostic models perform better than the SPL-MAD reference. \underline{Underlined} values are the best performing true one-class models.}
    \label{tab:one_class}

        \footnotesize{
            \renewcommand{\arraystretch}{1.2}
            \begin{tabular}{@{}lccccccc@{}}
                \toprule
                Src. dataset & \multicolumn{3}{c}{FRGC}  & \phantom{;}               & \multicolumn{3}{c}{FFHQ}                                                                                              \\
                \cmidrule{2-4}\cmidrule{6-8}
                Test attacks & LB                        & GAN                       & DIFF                      &                            & LB                        & GAN                       & DIFF \\
                \midrule
                Model        &                           &                           &                           &                            &                           &                           &      \\
                AIM          & \textbf{6.08}             & \textbf{\underline{0.39}} & \textbf{\underline{0.00}} & & 34.40                      & 56.10                     & \textbf{\underline{7.20}}        \\
                CLIP         & 23.87                     & \textbf{1.52}             & 20.92                     & & \textbf{\underline{14.50}} & \textbf{\underline{4.75}} & \textbf{27.70}                   \\
                DNADet       & \textbf{\underline{0.87}} & \textbf{0.82}             & \textbf{0.48}             & & \textbf{27.10}             & 29.10                     & \textbf{32.80}                   \\
                DINOv2       & 35.72                     & 32.86                     & 30.16                     & & 35.80                      & 48.90                     & \textbf{34.00}                   \\
                RN50-IN      & 51.56                     & 43.23                     & \textbf{18.75}            & & 46.75                      & 61.25                     & 46.10                            \\
                SPL-MAD *    & 16.28                     & 11.02                     & 20.23                     & & 28.15                      & 14.10                     & 34.20                            \\
                \bottomrule
            \end{tabular}
        }
    \end{threeparttable}

\end{table}

Finally, Table \ref{tab:one_class} presents the performance of one-class detectors trained only on bonafide data. It is important to note that the comparison to SPL-MAD is not entirely fair, as SPL-MAD is trained on Casia-Webface data, while our detectors are specifically tuned to the considered source dataset, providing an advantage. Nonetheless, for \ac{FRGC} attacks, AIM and DNADet probes show quite strong performance, and significantly better than SPL-MAD. DNADet probes in particular lead to an impressive \ac{D-EER} of under 1\% for all considered families of attacks, even though the detector is never exposed to any attack for its development. For \ac{FFHQ} attacks however, the overall detection performance is unsatisfactory, with CLIP features proving to be the most effective in this scenario.

\subsection{Discussion}
The results demonstrate that the considered attack-agnostic feature representations are highly effective for morphing attack detection. Training simple probes on these features consistently outperforms a CNN detector trained end-to-end on image samples across all generalization scenarios. They also lead to improved performance over an out-of-the-box one-class detector from the recent literature. However, \emph{which} representation is the most effective is scenario-dependent.

The key outcomes can be summarized as follows:

\begin{itemize}
\item \textbf{DNADet features} are particularly effective for one-class modeling in the digital domain and when targeting a single passport standard. The DNADet one-class detector achieves a \ac{D-EER} under 1\% for all attack families on FRGC attacks. However, these features exhibit poor performance in print-scan generalization. This limitation is likely due to DNADet’s pretraining task of contrastive learning of image transformations, which may result in a different representation manifold for print-scan images compared to digital ones. Incorporating print-scan data into the bonafide training set may resolve this issue, which we plan to explore in future work.

\item \textbf{AIM features} excel for generalizing to unseen attacks but show inconsistencies in other generalization scenarios. While AIM features behave overall similarly to DNADet features, their more irregular performance across different attack families may limit their practicality in real-world applications.

\item \textbf{DINOv2 features} are particularly suitable for print-scan generalization. In scenarios where we assume a limited known set of possible attacks (i.e., all attacks can be seen during training), these features are valuable when generating actual print-scan data for training is impractical or too time-consuming. Future work should in particular verify if this print-scan generalization performance holds across a wider variety of physical devices.

\item \textbf{CLIP features}, even though they are rarely the best, consistently perform well across all generalization scenarios, making them interesting for scenarios where multiple generalization challenges are simultaneous. By enabling robust generalization to unseen attacks, strong source dataset generalization, and decent print-scan generalization, they become a strong candidate for training detectors in a supervised way on a small set of attacks. In the one-class setting, CLIP features, while less effective than DNADet on FRGC attacks, are the most effective for FFHQ attacks. Coupled with their strong source dataset generalization capability, this fact makes them potentially well-suited for developing more general-purpose one-class \ac{MAD} systems that target \emph{multiple} passport standards.
\end{itemize}

\section{Conclusion}
Our work highlighted the superior effectiveness of training simple probes on top of attack-agnostic features in morphing attack detection (MAD) compared to traditional supervised CNN training (MixFaceNet) and a one-class detector from the literature (SPL-MAD).

In particular, DNADet features led to remarkable performance in one-class detection scenarios, achieving a \ac{D-EER} of less than 1\% for all attack families on the FRGC dataset. This underscores its efficacy in detecting morphs without prior exposure to attack samples. However, this performance was limited to the digital domain, with DNADet showing low efficacy for generalization to the print-scan domain. This indicates the need to explore whether the inclusion of bonafide print-scan data in the training set of the DNADet one-class model might enable similar performance in the print-scan domain.

Conversely, DINOv2 excelled in print-scan generalization, making it a promising candidate for contexts where generating large enough print-scan data for training is impractical.

Finally, CLIP, while not always the top performer, consistently delivered solid results across all generalization scenarios. This highlights its potential for developing more versatile MAD systems capable of handling various types of generalization.

Future work will focus on several key areas to further enhance the robustness and generalizability of our proposed approach. First, a more systematic evaluation of one-class detection performance is necessary, particularly to ensure fairer comparisons with existing methods, notably by making sure equivalent bonafide sets are seen at training time. Second, an evaluation of the one-class performance of DNADet in the print-scan domain is needed, likely requiring the inclusion of bonafide print-scan data in the training set. Third, there is potential in specializing attack-agnostic extractors by continuing pretraining using content-specific data, such as bonafide face images. In this work, only DNADet had been pretrained on face data. Lastly, the print-scan generalization capabilities of DINOv2 should be evaluated using additional print-scan devices to verify its effectiveness across a broader range of physical conditions.

In conclusion, the study validates the effectiveness of attack-agnostic representations for MAD, with DNADet and CLIP feature representations standing out in one-class and generalist performances, respectively, and DINOv2 in print-scan generalization. The outlined future work aims to address current limitations and further optimize these models for practical deployment in diverse real-world scenarios.

\section*{Acknowledgments}
This work was supported by the Swiss Center for Biometrics Research \& Testing and the Idiap Research Institute.

{\small
\bibliographystyle{ieee}
\bibliography{main.bib}

\begin{thebibliography}{10}\itemsep=-1pt

\bibitem{FaceAnalysisTechnology}
Face {{Analysis Technology Evaluation}} ({{FATE}}) {{MORPH}}.
\newblock https://pages.nist.gov/frvt/html/frvt\_morph.html.

\bibitem{FaceResearchLab2017}
Face {{Research Lab London Set}}, May 2017.

\bibitem{colboisDetectionMorphingAttacks2022}
L.~Colbois and S.~Marcel.
\newblock On the detection of morphing attacks generated by {{GANs}}.
\newblock In {\em 2022 {{International Conference}} of the {{Biometrics Special
  Interest Group}} ({{BIOSIG}})}, pages 1--5, Sept. 2022.

\bibitem{cozzolinoRaisingBarAIgenerated2023}
D.~Cozzolino, G.~Poggi, R.~Corvi, M.~Nie{\ss}ner, and L.~Verdoliva.
\newblock Raising the {{Bar}} of {{AI-generated Image Detection}} with
  {{CLIP}}, Nov. 2023.

\bibitem{damerMorDIFFRecognitionVulnerability2023}
N.~Damer, M.~Fang, P.~Siebke, J.~N. Kolf, M.~Huber, and F.~Boutros.
\newblock {{MorDIFF}}: {{Recognition Vulnerability}} and {{Attack
  Detectability}} of {{Face Morphing Attacks Created}} by {{Diffusion
  Autoencoders}}.
\newblock In {\em 2023 11th {{International Workshop}} on {{Biometrics}} and
  {{Forensics}} ({{IWBF}})}, pages 1--6, Apr. 2023.

\bibitem{damerPrivacyfriendlySyntheticData2022a}
N.~Damer, C.~A.~F. L{\'o}pez, M.~Fang, N.~Spiller, M.~V. Pham, and F.~Boutros.
\newblock Privacy-friendly {{Synthetic Data}} for the {{Development}} of {{Face
  Morphing Attack Detectors}}.
\newblock In {\em 2022 {{IEEE}}/{{CVF Conference}} on {{Computer Vision}} and
  {{Pattern Recognition Workshops}} ({{CVPRW}})}, pages 1605--1616, June 2022.

\bibitem{el-noubyScalablePretrainingLarge2024}
A.~{El-Nouby}, M.~Klein, S.~Zhai, M.~A. Bautista, A.~Toshev, V.~Shankar, J.~M.
  Susskind, and A.~Joulin.
\newblock Scalable {{Pre-training}} of {{Large Autoregressive Image Models}},
  Jan. 2024.

\bibitem{fangUnsupervisedFaceMorphing2022}
M.~Fang, F.~Boutros, and N.~Damer.
\newblock Unsupervised {{Face Morphing Attack Detection}} via {{Self-paced
  Anomaly Detection}}, Aug. 2022.

\bibitem{ferraraMagicPassport2014}
M.~Ferrara, A.~Franco, and D.~Maltoni.
\newblock The magic passport.
\newblock In {\em {{IEEE International Joint Conference}} on {{Biometrics}}},
  pages 1--7, Sept. 2014.

\bibitem{ivanovskaFaceMorphingAttack2023}
M.~Ivanovska and V.~{\v S}truc.
\newblock Face {{Morphing Attack Detection}} with {{Denoising Diffusion
  Probabilistic Models}}.
\newblock {\em 2023 11th International Workshop on Biometrics and Forensics
  (IWBF)}, pages 1--6, Apr. 2023.

\bibitem{karrasStyleBasedGeneratorArchitecture2019}
T.~Karras, S.~Laine, and T.~Aila.
\newblock A {{Style-Based Generator Architecture}} for {{Generative Adversarial
  Networks}}.
\newblock In {\em 2019 {{IEEE}}/{{CVF Conference}} on {{Computer Vision}} and
  {{Pattern Recognition}} ({{CVPR}})}, pages 4396--4405, June 2019.

\bibitem{makrushinAutomaticGenerationDetection2017}
A.~Makrushin, T.~Neubert, and J.~Dittmann.
\newblock Automatic {{Generation}} and {{Detection}} of {{Visually Faultless
  Facial Morphs}}:.
\newblock In {\em Proceedings of the 12th {{International Joint Conference}} on
  {{Computer Vision}}, {{Imaging}} and {{Computer Graphics Theory}} and
  {{Applications}}}, pages 39--50, Porto, Portugal, 2017. {SCITEPRESS - Science
  and Technology Publications}.

\bibitem{neubertExtendedStirTraceBenchmarking2018}
T.~Neubert, A.~Makrushin, M.~Hildebrandt, C.~Kr{\"a}tzer, and J.~Dittmann.
\newblock Extended {{StirTrace}} benchmarking of biometric and forensic
  qualities of morphed face images.
\newblock {\em IET Biom.}, 2018.

\bibitem{ojhaUniversalFakeImage2023}
U.~Ojha, Y.~Li, and Y.~J. Lee.
\newblock Towards {{Universal Fake Image Detectors}} that {{Generalize Across
  Generative Models}}.
\newblock In {\em 2023 {{IEEE}}/{{CVF Conference}} on {{Computer Vision}} and
  {{Pattern Recognition}} ({{CVPR}})}, pages 24480--24489, Vancouver, BC,
  Canada, June 2023. IEEE.

\bibitem{oquabDINOv2LearningRobust2023}
M.~Oquab, T.~Darcet, T.~Moutakanni, H.~Vo, M.~Szafraniec, V.~Khalidov,
  P.~Fernandez, D.~Haziza, F.~Massa, A.~{El-Nouby}, M.~Assran, N.~Ballas,
  W.~Galuba, R.~Howes, P.-Y. Huang, S.-W. Li, I.~Misra, M.~Rabbat, V.~Sharma,
  G.~Synnaeve, H.~Xu, H.~Jegou, J.~Mairal, P.~Labatut, A.~Joulin, and
  P.~Bojanowski.
\newblock {{DINOv2}}: {{Learning Robust Visual Features}} without
  {{Supervision}}, Apr. 2023.

\bibitem{paszkePyTorchImperativeStyle2019}
A.~Paszke, S.~Gross, F.~Massa, A.~Lerer, J.~Bradbury, G.~Chanan, T.~Killeen,
  Z.~Lin, N.~Gimelshein, L.~Antiga, A.~Desmaison, A.~Kopf, E.~Yang, Z.~DeVito,
  M.~Raison, A.~Tejani, S.~Chilamkurthy, B.~Steiner, L.~Fang, J.~Bai, and
  S.~Chintala.
\newblock {{PyTorch}}: {{An Imperative Style}}, {{High-Performance Deep
  Learning Library}}.
\newblock In {\em Advances in {{Neural Information Processing Systems}}},
  volume~32. Curran Associates, Inc., 2019.

\bibitem{phillipsOverviewFaceRecognition2005}
P.~Phillips, P.~Flynn, T.~Scruggs, K.~Bowyer, J.~Chang, K.~Hoffman, J.~Marques,
  J.~Min, and W.~Worek.
\newblock Overview of the face recognition grand challenge.
\newblock In {\em 2005 {{IEEE Computer Society Conference}} on {{Computer
  Vision}} and {{Pattern Recognition}} ({{CVPR}}'05)}, volume~1, pages 947--954
  vol. 1, June 2005.

\bibitem{radfordLearningTransferableVisual2021}
A.~Radford, J.~W. Kim, C.~Hallacy, A.~Ramesh, G.~Goh, S.~Agarwal, G.~Sastry,
  A.~Askell, P.~Mishkin, J.~Clark, G.~Krueger, and I.~Sutskever.
\newblock Learning {{Transferable Visual Models From Natural Language
  Supervision}}.
\newblock In {\em Proceedings of the 38th {{International Conference}} on
  {{Machine Learning}}}, pages 8748--8763. PMLR, July 2021.

\bibitem{rajaMorphingAttackDetectionDatabase2021}
K.~Raja, M.~Ferrara, A.~Franco, L.~Spreeuwers, I.~Batskos, F.~De~Wit,
  M.~{Gomez-Barrero}, U.~Scherhag, D.~Fischer, S.~K. Venkatesh, J.~M. Singh,
  G.~Li, L.~Bergeron, S.~Isadskiy, R.~Ramachandra, C.~Rathgeb, D.~Frings,
  U.~Seidel, F.~Knopjes, R.~Veldhuis, D.~Maltoni, and C.~Busch.
\newblock Morphing {{Attack Detection-Database}}, {{Evaluation Platform}}, and
  {{Benchmarking}}.
\newblock {\em IEEE Transactions on Information Forensics and Security},
  16:4336--4351, 2021.

\bibitem{sarkarAreGANbasedMorphs2022}
E.~Sarkar, P.~Korshunov, L.~Colbois, and S.~Marcel.
\newblock Are {{GAN-based}} morphs threatening face recognition?
\newblock In {\em {{ICASSP}} 2022 - 2022 {{IEEE International Conference}} on
  {{Acoustics}}, {{Speech}} and {{Signal Processing}} ({{ICASSP}})}, pages
  2959--2963, May 2022.

\bibitem{scherhagFaceMorphingAttack2022}
U.~Scherhag, C.~Rathgeb, and C.~Busch.
\newblock Face {{Morphing Attack Detection Methods}}.
\newblock In C.~Rathgeb, R.~Tolosana, R.~{Vera-Rodriguez}, and C.~Busch,
  editors, {\em Handbook of {{Digital Face Manipulation}} and {{Detection}}:
  {{From DeepFakes}} to {{Morphing Attacks}}}, pages 331--349. Springer
  International Publishing, Cham, 2022.

\bibitem{tapiaFaceFeatureVisualisation2023}
J.~E. Tapia and C.~Busch.
\newblock Face {{Feature Visualisation}} of {{Single Morphing Attack
  Detection}}.
\newblock {\em 2023 11th International Workshop on Biometrics and Forensics
  (IWBF)}, pages 1--6, Apr. 2023.

\bibitem{venkateshCanGANGenerated2020}
S.~Venkatesh, H.~Zhang, R.~Ramachandra, K.~Raja, N.~Damer, and C.~Busch.
\newblock Can {{GAN Generated Morphs Threaten Face Recognition Systems
  Equally}} as {{Landmark Based Morphs}}? - {{Vulnerability}} and
  {{Detection}}.
\newblock In {\em 2020 8th {{International Workshop}} on {{Biometrics}} and
  {{Forensics}} ({{IWBF}})}, pages 1--6, Apr. 2020.

\bibitem{yangDeepfakeNetworkArchitecture2022a}
T.~Yang, Z.~Huang, J.~Cao, L.~Li, and X.~Li.
\newblock Deepfake {{Network Architecture Attribution}}.
\newblock {\em Proceedings of the AAAI Conference on Artificial Intelligence},
  36(4):4662--4670, June 2022.

\bibitem{zhangMIPGANGeneratingStrong2021}
H.~Zhang, S.~Venkatesh, R.~Ramachandra, K.~Raja, N.~Damer, and C.~Busch.
\newblock {{MIPGAN}}---{{Generating Strong}} and {{High Quality Morphing
  Attacks Using Identity Prior Driven GAN}}.
\newblock {\em IEEE Transactions on Biometrics, Behavior, and Identity
  Science}, 3(3):365--383, July 2021.

\end{thebibliography}
}




\end{document}